\documentclass[sigconf]{acmart}

\usepackage{booktabs}
\usepackage{caption}
\usepackage{subcaption}
\usepackage{multirow}
\usepackage{sg-ibs-macros} 

\usepackage{color}
\definecolor{darkred}{rgb}{0.7,0.1,0.1}
\definecolor{darkgreen}{rgb}{0.1,0.7,0.1}
\definecolor{cyan}{rgb}{0.7,0.0,0.7}
\definecolor{dblue}{rgb}{0.2,0.2,0.8}
\definecolor{maroon}{rgb}{0.76,.13,.28}
\definecolor{burntorange}{rgb}{0.81,.33,0}
\usepackage{xcolor}

\usepackage[ruled]{algorithm2e}

\SetAlFnt{\small}
\SetAlCapFnt{\small}
\SetAlCapNameFnt{\small}
\SetAlCapHSkip{0pt}

\copyrightyear{2022} 
\acmYear{2022} 
\setcopyright{acmcopyright}
\acmConference[SIGGRAPH '22 Conference Proceedings]{Special Interest Group on Computer Graphics and Interactive Techniques Conference Proceedings}{August 7--11, 2022}{Vancouver, BC, Canada}
\acmBooktitle{Special Interest Group on Computer Graphics and Interactive Techniques Conference Proceedings (SIGGRAPH '22 Conference Proceedings), August 7--11, 2022, Vancouver, BC, Canada}
\acmPrice{15.00}
\acmDOI{10.1145/3528233.3530710}
\acmISBN{978-1-4503-9337-9/22/08}


\begin{document}

\begin{teaserfigure}
\centering
\begin{tabular}{cccc}
\includegraphics[width=0.22\textwidth]{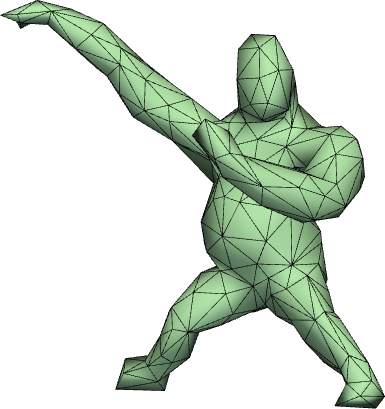}&
\includegraphics[width=0.22\textwidth]{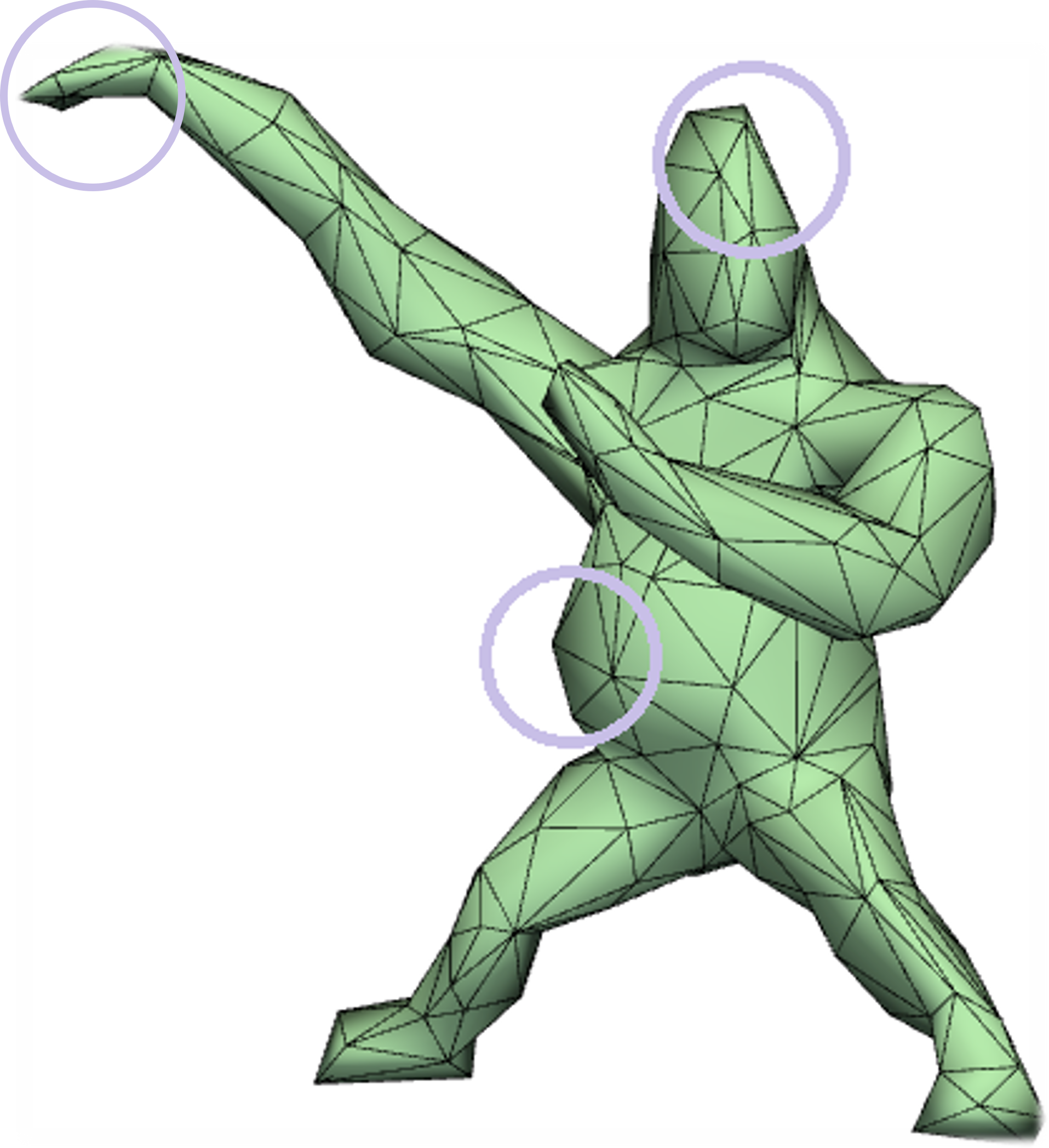}&
\includegraphics[width=0.22\textwidth]{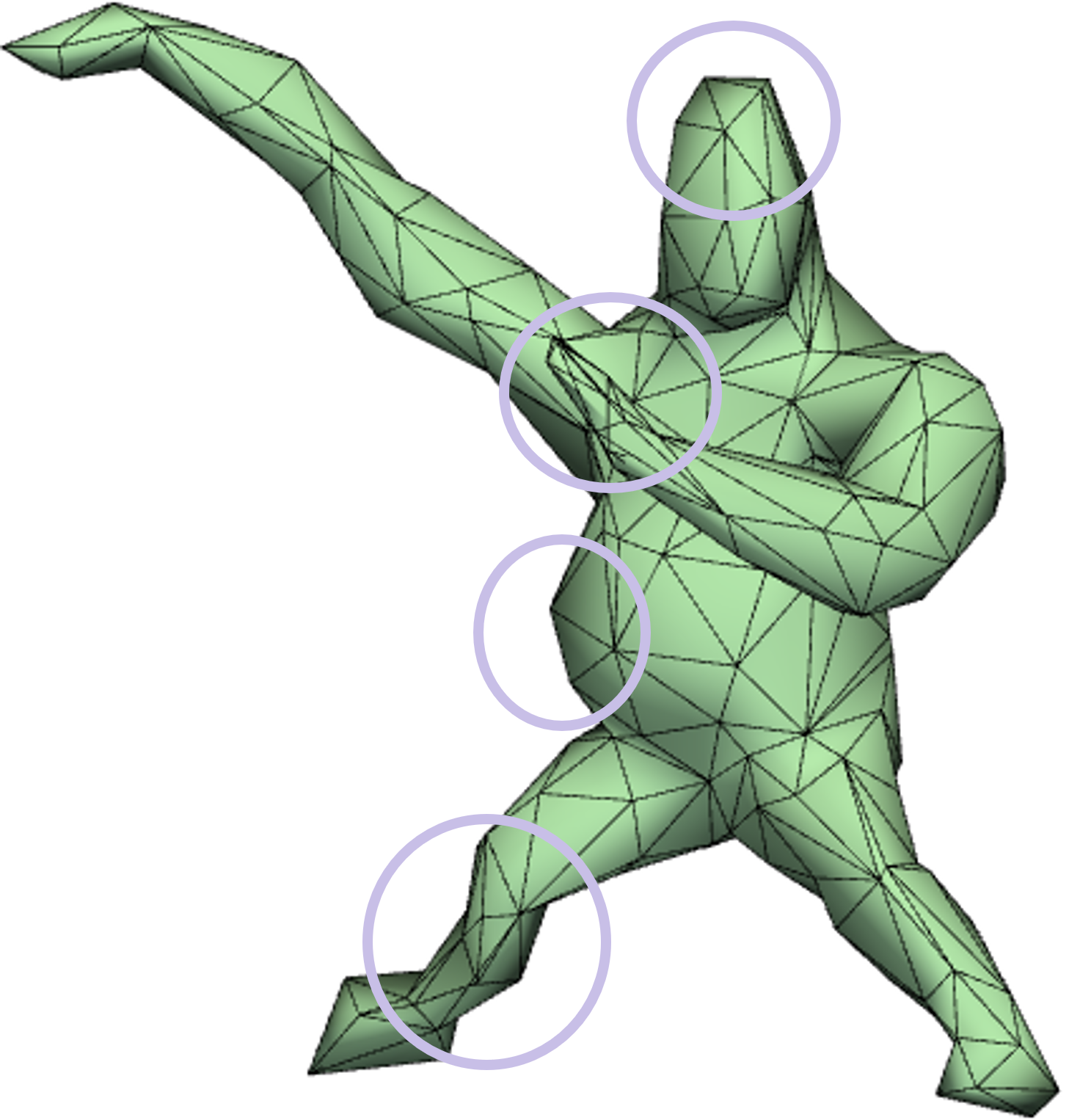}&
\includegraphics[width=0.22\textwidth]{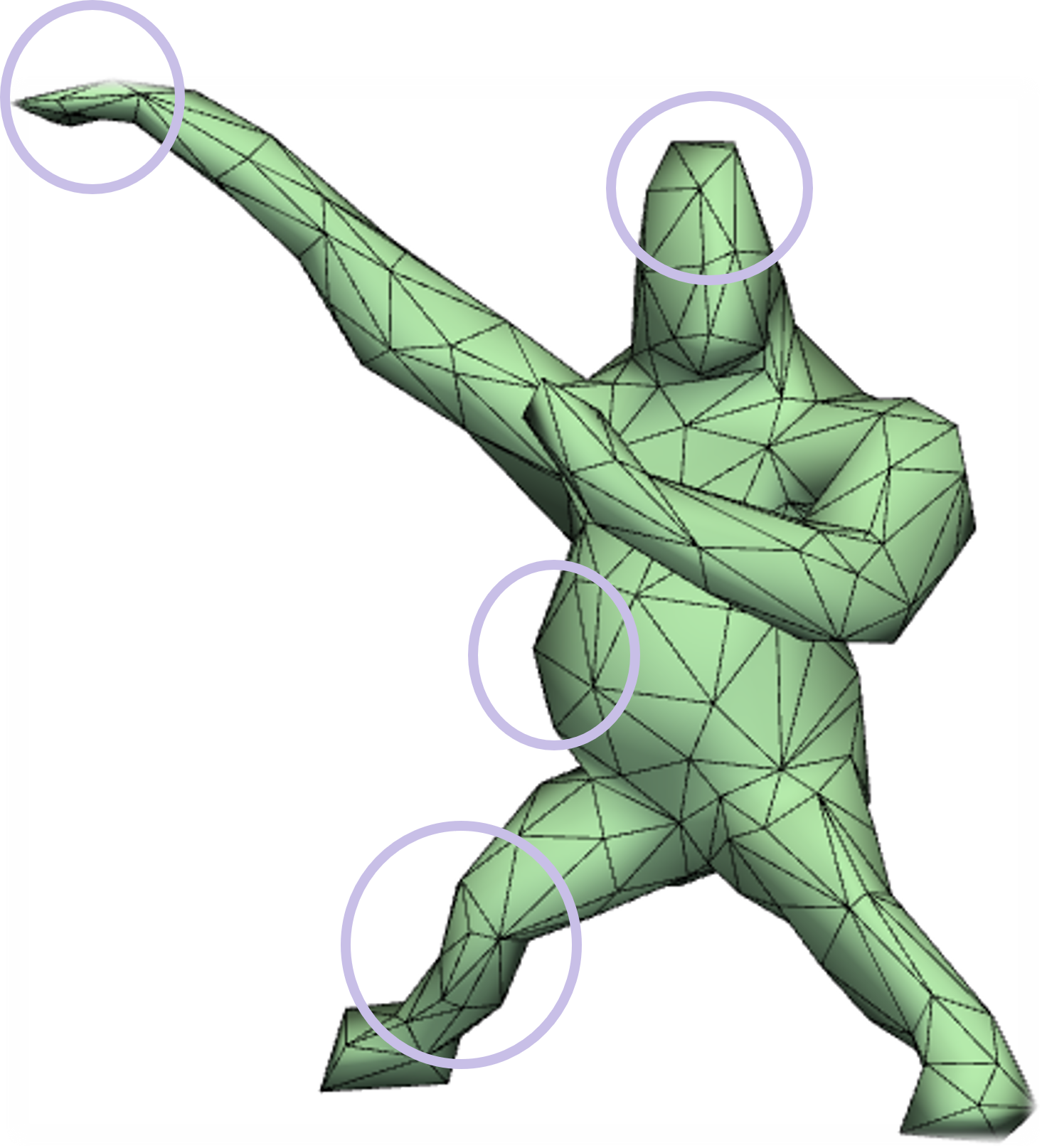}\\
(a) Input &
(b) MeshWalker 
&
(c) PD-MeshNet 
&
(d) MeshCNN 
\\
Class: {\color{darkgreen}{Gorilla}} &
Classified: {\color{darkred}{Alien}} &
Classified: {\color{darkred}{Horse}} &
Classified: {\color{darkred}{Armadillo}}
\end{tabular}
\caption{
{\bf Adversarial meshes.} 
(a) Given a mesh of  a gorilla,
it is attacked by modifying the mesh regions proportionally to the influence they have on the specific SOTA classification  network~\cite{lahav2020meshwalker,Milano20NeurIPS-PDMeshNet,hanocka2019meshcnn}.  
The most  modified regions are marked by gray circles.
 Although the modifications are barely visible, all the networks are misled and misclassify the gorilla as either an alien, a horse or an armadillo.
 }
 \label{fig:teaser}
\end{teaserfigure}

\title{Random Walks for Adversarial Meshes}
\author{Amir Belder}
\affiliation{%
  \institution{Technion – Israel Institute of Technology}
  \country{Israel}
}
\email{amirbelder@campus.technion.ac.il}

\author{Gal Yefet}
\affiliation{%
  \institution{Technion – Israel Institute of Technology}
  \country{Israel}
}
\email{galyefet@campus.technion.ac.il}

\author{Ran Ben Izhak}
\affiliation{%
  \institution{Technion – Israel Institute of Technology}
  \country{Israel}
}
\email{benizhakran@gmail.com}

\author{Ayellet Tal}
\affiliation{%
  \institution{Technion – Israel Institute of Technology}
  \country{Israel}
}
\email{ayellet@ee.technion.ac.il}


\begin{abstract}
A polygonal mesh is the most-commonly used representation of surfaces in computer graphics.
Therefore, it is not surprising that a number of mesh classification networks have recently been proposed. 
However, while adversarial attacks are wildly researched in 2D, the field of adversarial meshes is under explored.
This paper proposes a novel, unified, and general adversarial attack, which leads to misclassification of 
several state-of-the-art mesh classification neural networks.
Our attack approach is black-box, i.e. it has access only to the network's predictions, but not to the network's full architecture or gradients.
The key idea is to train a network to imitate a given classification network.
This is done by utilizing random walks along the mesh surface, which gather geometric information. 
These walks provide insight onto the regions of the mesh that are important for the correct prediction of the given classification network.
These mesh regions are then modified more than other regions in order to attack the network in a manner that is barely visible to the naked eye.
\end{abstract}


\maketitle
\section{Introduction}

Neural networks achieve outstanding results in numerous tasks in computer vision \& graphics,
however they are oftentimes  vulnerable to adversarial attacks ~\cite{carlini2017towards,DBLP:journals/corr/abs-1708-06131, szegedy2013intriguing, madry2017towards, wallace2019universal, ebrahimi2017hotflip, carlini2018audio,NEURIPS2020_b8ce4776, Croce_2019_ICCV, Rony_2019_CVPR, Xie_2019_CVPR,papernot2017practical, tsai2020robust}.
These attacks modify the input data in a way that is hardly visible to the naked eye,  yet leads to misclassification.
Adversarial attacks can be divided into categories: {\em white-box}, {\em black-box}, and {\em gray-box}.
In white-box, the attacker has access to the networks' full architecture, gradients and predictions~\cite{goodfellow2014explaining,DBLP:journals/corr/PapernotMGJCS16}.
In black-box, the attacker has access only to the networks' predictions~\cite{DBLP:journals/corr/PapernotMGJCS16}.
In gray-box, 
the attacker has access to more than just the predictions, but not to the full architecture~\cite{Vivek_2018_ECCV}.

This paper focuses on three-dimensional data, differently from the works mentioned above that attack images. 
In 3D, most works attempt to attack point-cloud networks~\cite{liu2019extending,xiang2019generating,10.1007/978-3-030-58610-2_15,Zhao_2020_CVPR,Wicker_2019_CVPR,DBLP:journals/corr/abs-2104-12146}.
The key idea is to move specific points, whose displacement will lead to misclassification. 
Since point clouds have no topological constraints,  such movements would be hardly visible.
Polygonal meshes, however, are inherently different in that aspect.
Moving even a single vertex might cause a highly noticeable movement of the adjacent edges \& faces and possibly result in topological changes such as self-intersections.
If the meshes are textured 
the attacks may change the texture of the meshes to cause misclassification~\cite{yao2020multiview,yang2018realistic,Xiao_2019_CVPR}.
There are also works that attack graph networks~\cite{zhang2019data, sun2018data, chen2017practical}.
These attacks modify the graph structure, which is undesirable for meshes.

We address the task of attacking the most popular representation of 3D data---texture-less meshes.
As this representation is used in many applications, including modeling, animation and medical purposes, acknowledging the vulnerabilities of mesh classification networks is important.
For meshes, white-box attacks were proposed~\cite{rampini2021universal, mariani2020generating, DBLP:journals/corr/abs-2104-12146}.

We propose a novel, black-box, unified, and general adversarial attack, which leads to misclassification in SOTA mesh neural networks, as shown in Fig~\ref{fig:teaser}.
At the base of our method lies the concept of an {\em imitating network}. 
We propose to train  a network to {\em imitate} a given classification network.
For each network we wish to attack, our imitating network gets as input pairs of a mesh \& a prediction vector for that mesh (i.e. querying all meshes in the dataset).
It basically learns the classification function of the given attacked network, by learning to generate prediction vectors for the meshes.
For this to be done, our loss function should consider the distribution of the prediction vectors rather than one-hot label vectors used for classification.

Our imitating network utilizes random walks along the mesh.
As shown in ~\cite{lahav2020meshwalker},
random walks are a powerful tool for mesh exploration, gathering both global and local geometric information on the object.
During a walk, features of various importance are extracted, as obviously some regions of the mesh are more distinctive than others.
The state-of-the-art classification networks inherently differ from each other, both in the manner they extract their features and in the parts of the meshes that they focus on~\cite{feng2019meshnet,hanocka2019meshcnn, lahav2020meshwalker,Milano20NeurIPS-PDMeshNet}.
Therefore, architecture-dependent changes are needed in order to cause each of the SOTA systems to misclassify.   
We will show that our random walk-based network manages to learn these specific architecture-dependent features. 

Our objective, however, is not only to cause misclassification, but also to do so while minimizing the change to the input meshes. 
If our modifications focus solely on the distinct regions, the network is going to be misled indeed, but the attacked meshes will look bad.
For instance, a camel with no hump is not a distinctive camel. 
Thus, the two goals of the attack---misclassification and remaining distinct---might contradict.
We will show that random walks suit these contradicting goals: 
As they wander around the surface, the modifications are spread across the entire mesh, both in distinctive and in non-distinctive regions.
The need to spread the modifications is enhanced by \cite{Ben_Izhak_2022_WACV}'s observation that all vertices contribute to the classification, though to varying degrees.
In our method,  as a vertex's influence on classification grows, so does its modification.
This influence is manifested in the gradient of the classification loss. 
Therefore, we change the mesh vertices in the opposite direction of the gradients.

Our approach is shown to fool  SOTA classification networks on two of the most commonly-used datasets.
For each dataset, our approach attacks the networks that report the best results for that dataset. 
In particular, we attack  (1)  MeshCNN~\cite{hanocka2019meshcnn}, PD-MeshNet~\cite{Milano20NeurIPS-PDMeshNet} and MeshWalker~\cite{lahav2020meshwalker} on SHREC11~\cite{veltkamp2011shrec} and (2) MeshNet~\cite{feng2019meshnet} and MeshWalker~\cite{lahav2020meshwalker} on ModelNet40~\cite{wu20153d}.
These datasets are chosen not only due to their prevalence, but also because the SOTA networks achieve excellent results for them, which is important in order to verify that excellent classifiers can be misled.
As a typical example, PD-MeshNet achieves $99.7\%$ accuracy before the attack and $18.3\%$ after it on SHREC11. 

Fig.~\ref{fig:teaser} illustrates how very small modifications to the original mesh, which would seem insignificant to a person, cause different SOTA systems to misclassify the object.
Moreover, the figure shows how different parts of the mesh are changed in order to mislead each of the networks:
The right hand of the Gorilla is modified in order to fool MeshWalker, the left hand and the lower leg to mislead PD-MeshNet, and the right hand and the lower leg to fool MeshCNN; interestingly, the head and the belly are modified in all cases.
This is an additional benefit of our network---it detects the parts of the object each of these networks focuses on for classification.
This provides a rare opportunity to shed some light on how these networks classify.

This paper makes the following contributions:
First, 
    it presents a novel, unified and general black-box approach to attack mesh classification networks.
    Furthermore, it proposes a network that realizes this approach.
Second,
     it demonstrates how vulnerable to adversarial attacks the current SOTA classification networks are. 
     This can be useful for developing more robust networks.
     Moreover, profound insight is gained onto the parts of the mesh that are important for the correct prediction of the different networks.

\section{Method}

\begin{figure*}[tb]
\centering
\includegraphics[width=0.95\textwidth]{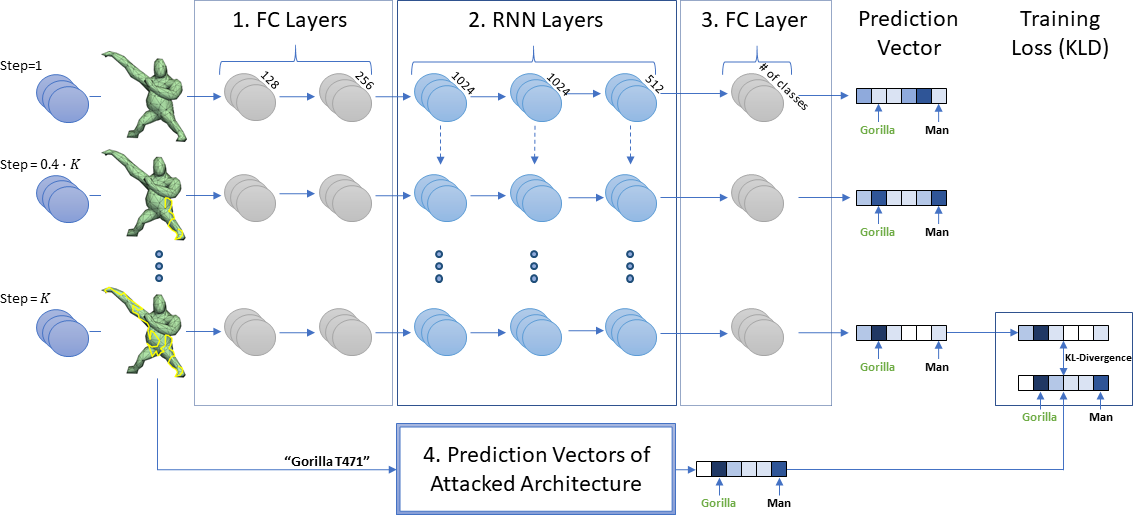}
\caption{
{\bf Imitating network architecture.} 
Our network gets a mesh, a random walk (the yellow walk on the gorilla) and a prediction vector.
As in~\cite{lahav2020meshwalker}, it consists of $3$ components: 
(1) The FC layers change the feature space; 
(2) The RNN layers aggregate the information along the walk; 
(3) The  FC layer predicts the outcome of the network. 
The KLD loss is applied to the prediction of the last walk vertex jointly with the imitated network prediction.
}
\label{fig:imitating}
\end{figure*}

We are given a dataset of meshes, partitioned into categories, and
all the prediction vectors of the classification network to be attacked.  
Adversarial attacks aim to change the input in a manner that would cause the network to misclassify the input, while a person would still classify it as its true  class.
In the following we propose an approach that, 
when given a mesh, will generate such a modified mesh. 

The basic question is where on the mesh these changes shall be and how they shall be spread.
Evidently, distinctive regions must be modified in order to cause misclassification.
However, if the modifications are concentrated on these regions, they are likely to be noticeable to the naked eye, which will not satisfy our constraint.
This creates a delicate equilibrium that must be upheld for a successful adversarial attack: spreading the change across the mesh, while still fooling the classification network.

Our key idea is to randomly walk along the mesh edges and slightly move the vertices along the walk.
A {\em random walk} is extracted by randomly choosing an initial vertex, and then iteratively adding  vertices to the sequence from the adjacent vertices  to the last vertex of the sequence (if it is not already there).
We draw inspiration from the MeshWalker network~\cite{lahav2020meshwalker}, which shows that  random walks are a powerful tool for classification.
At the base of our method lies the observation that some portions of each walk influence the classification more than others. 
Furthermore, the amount of influence is manifested in the gradient of the classification loss.
The more influential a vertex is , the larger its gradient will be.
Thus, our attack is gradient-based:
By changing the vertices against the gradients, the most influential vertices will change the most, thereby changing the mesh in a manner that makes it less likely to be classified correctly.
However, as all the vertices of the walk contribute to the classification, all of them will change during the attack, hence spreading the changes also to non-distinctive areas.
This is crucial in achieving the above target equilibrium. 

Our attack approach is general and unified, i.e. it can be applied to any mesh classification network.
In particular, for a given classification network we wish to attack, we train an \underline{{\em imitating network}}.
Given meshes and their corresponding prediction vectors, queried from the network it shall attack,
the imitating network learns  similar attributes to those of the imitated network.
We elaborate below.

\subsection{Training}
Each imitating network gets as input: (1)~the train meshes of a dataset and (2)~
the prediction vectors of the attacked network on these meshes, i.e. the full probability vectors of  the dataset classes.
The goal is to learn to imitate the prediction function, so as to learn the same traits as the network it is imitating.
We note that each classification network preforms a different type of non-trainable pre-processing to the meshes before classifying them, which usually includes mesh simplification.
We preform the same pre-processing.

Regardless of the network we wish to attack, our imitating network has the architecture illustrated in Fig.~\ref{fig:imitating}.
It is similar to the architecture  of MeshWalker~\cite{lahav2020meshwalker}, except for two differences: the input and the loss function.
Given a random walk that consists of vertices (3D coordinates) along the mesh, referred to as {\em steps}, the data is aggregated as follows. 
First, a couple of FC layers upscale each vertex into $256$ dimensions. 
The second component is a {\em{recurrent neural network (RNN)}}, whose defining property is being able to "remember" and accumulate knowledge.
Thus, it aggregates the information of the vertices, "remembering" the walk's history.
A final FC layer predicts the classification.
The objective of our imitating network differs from that of~\cite{lahav2020meshwalker}:
Rather than predicting the classes of the meshes,
our imitating network drives to imitate a specific function---the prediction function of the imitated network.
Therefore, rather than requiring one-hot vectors of the source classes of the meshes as input, the imitating network requires the full probabilities vectors of the system it is imitating.
This is reflected in the type of the loss function used for training.

\begin{figure*}[htb]
\centering
\includegraphics[width=0.95\textwidth]{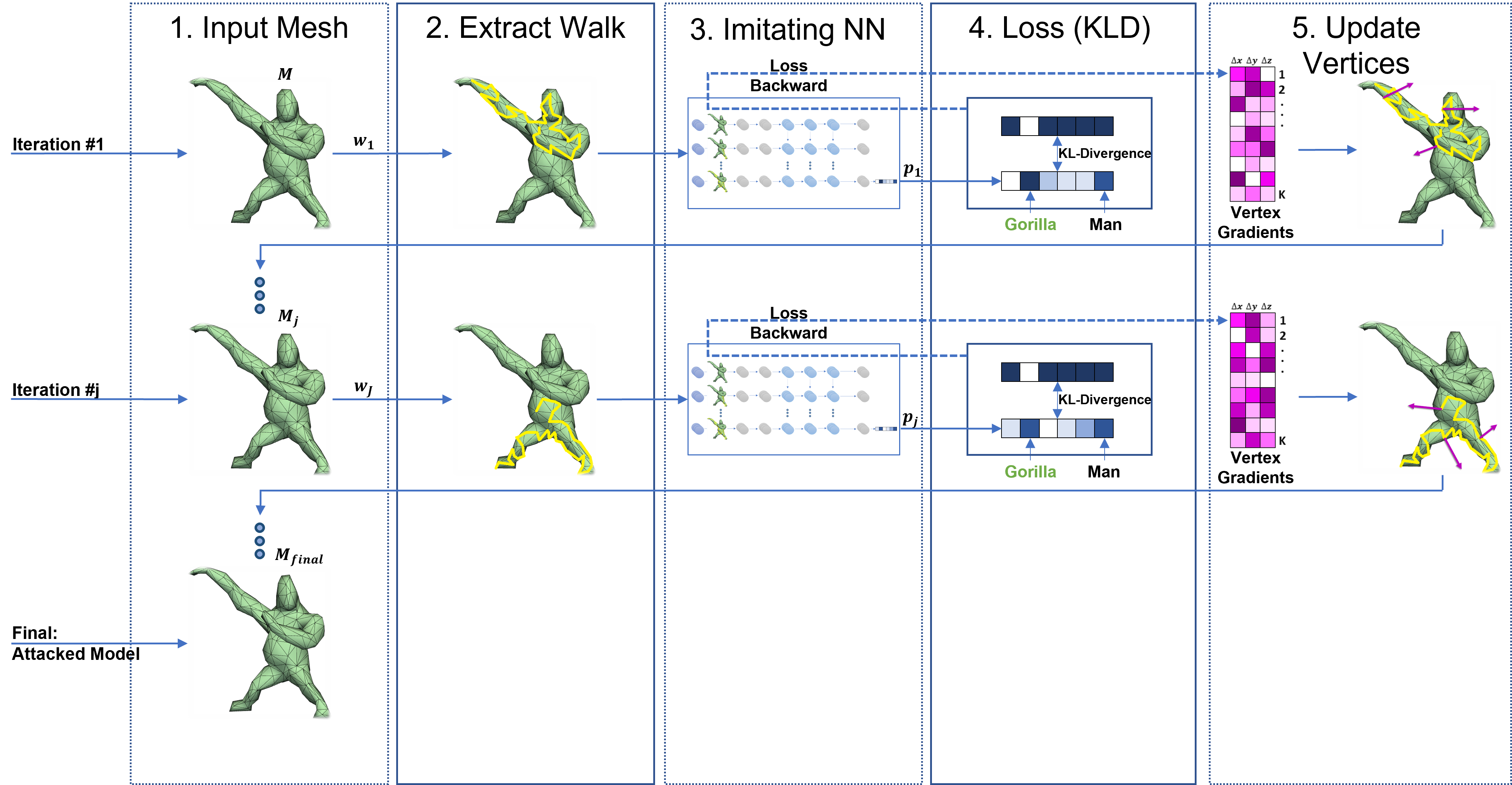}
\caption{
{\bf Attack.} 
At every iteration, a walk is extracted and our imitating network is applied to this walk, producing a  prediction vector. 
The gradients are computed for each vertex along the walk and the vertex location is updated w.r.t its gradient.
The modified mesh is fed into the next iteration, with a new walk.
}
\label{fig:architecture}
\end{figure*}

We use {\em Kullback–Leibler divergence (KLD)} loss function, instead of the sparse cross-entropy of~\cite{lahav2020meshwalker}, as we aim measure a distance between probabilities. 
Let $P_\theta$ be the posterior probability for observation $O$, computed by the trained network, parameterized with $\theta$,
$P_{ref}$ be the given ground truth probability for $O$,
$t_i$ (/ $r_i$) be the probability that the input mesh belongs to class $i$ out of $D$  classes in the dataset.
The loss function is defined as:
\begin{equation} 
KLD (P_\theta(t),\ P_{\mathrm{r}\mathrm{e}\mathrm{f}}(r))=
\sum^{D}_{i=1} p_{\mathrm{r}\mathrm{e}\mathrm{f}}(r_{i})\log\frac{p_{\mathrm{r}\mathrm{e}\mathrm{f}}(r_{i})}{p_{\theta}(t_{i})} .
\label{eq:KLD}
\end{equation}

\subsection{Attack.}
Once an imitating network $\mathcal{N}$ of a given mesh classification network is trained, the latter can be  attacked by  $\mathcal{N}$, as described in Algorithm~\ref{alg:MeshWalkerTraining}.
Given a  mesh $\mathcal{M}$ to be attacked, the following process iterates until $\mathcal{N}$ predicts the class of $\mathcal{M}$ as something other than its given class:
First, a random walk $w_{j}$ is extracted and
$\mathcal{N}$ produces its prediction for $w_{j}$, $p_{j}$.
If the predicted class differs from $\mathcal{M}$'s true class, $\mathcal{M}$ is saved as the attacked model, i.e. this is the stopping condition (in practice, we exit when misclassification occurs a few times).
Otherwise, the $KLD$ Loss between the prediction and the given one-hot label vector is calculated.
The gradients of the walk's vertices are calculated according to the result of the loss, as explained below.
Finally, the walk vertices are changed in the opposite direction of their gradients.
In effect,  the vertices of the walk that contribute more than others to the classification will change more.
Every additional walk modifies the previous slightly-modified mesh.
Fig.~\ref{fig:architecture} illustrates our attack. 

\begin{algorithm}[t]
\SetAlgoNoLine
\KwIn{A mesh $\mathcal{M}$, a one-hot label vector (of $\mathcal{M}$) $r$,
an imitating network $\mathcal{N}$}
\KwOut{Attacked model $\mathcal{M} (final)$}
\For{$j\gets 1$ \KwTo MAX\_ITERATIONS}{
    ${w_{j} \gets ExtractWalk(M)}$\;
    ${p}_{j} \gets Prediction(N, w_{j})$\;
    \If{$Softmax({p}_{j}) != r:$}
    {
        $Return$ $ M$ (final)\;
    }
    ${loss} \gets KLD(p_{j}, r)$\;
    ${gradients} \gets CalcGradients(loss, w_{j})$\;
    ${M} \gets UpdateMesh(M, gradients)$\;
    }
\caption{Adversarial Attacks via Random Walks}
\label{alg:MeshWalkerTraining}
\end{algorithm}

Specifically, our objective is to update the input mesh according to the network's loss function and prediction.
This is somewhat similar to the objective of back-propagation during training, where the weights of the network are updated according to the effect they had over the network's loss function and prediction.
In both cases, the influence is manifested in the derivative of the loss function.
Thus, we wish to measure how each vertex along the walk influences the loss and to change the vertex's coordinates accordingly, in order to cause misclassification.
Recall that our trained imitating network is {\em{RNN}}-based.
Thus, it aggregates the data of the walk, vertex by vertex, and "remembers" the history of the walk.
The network's memory of each step along the walk is manifested in its {\em state}, which is updated
after every step. 
(The last state is also used for prediction.)
Neural networks compute the gradients backwards for every layer, all the way, up to the input layer of the model (the trainable parameters in adversarial attacks).
Therefore, we get the {\em{gradient}} for each vertex, which is a 3D vector, i.e. the size of the input we started with.

In the attack's last stage, we modify the coordinates of a vertex according to the above gradient, by moving the vertex by $(\Delta x,\Delta y,\Delta z)$, that is computed by multiplying the gradient by a small constant, $\alpha$:
$(\Delta x,\Delta y,\Delta z)={\mathrm{\alpha}} * gradient$.
$\alpha$ is needed in order to reduce the likelihood of  self-intersections when moving the vertices. 
In practice $\alpha= 0.01$ for normalized datasets. 

In Section~\ref{sec:experiments} we will show that our attacks indeed achieve the two main goals: misclassification and visual resemblance to the original mesh.
An additional benefit of our attack, demonstrated in Fig.~\ref{fig:teaser}, is that it provides a rare opportunity to study which parts of the mesh are important to each classification system, i.e. the changes it is vulnerable to.
We will discuss this as well in Section~\ref{sec:experiments}.

\begin{figure*}[h]
\centering
\begin{tabular}{cccccc}
\includegraphics[width=0.18\textwidth]{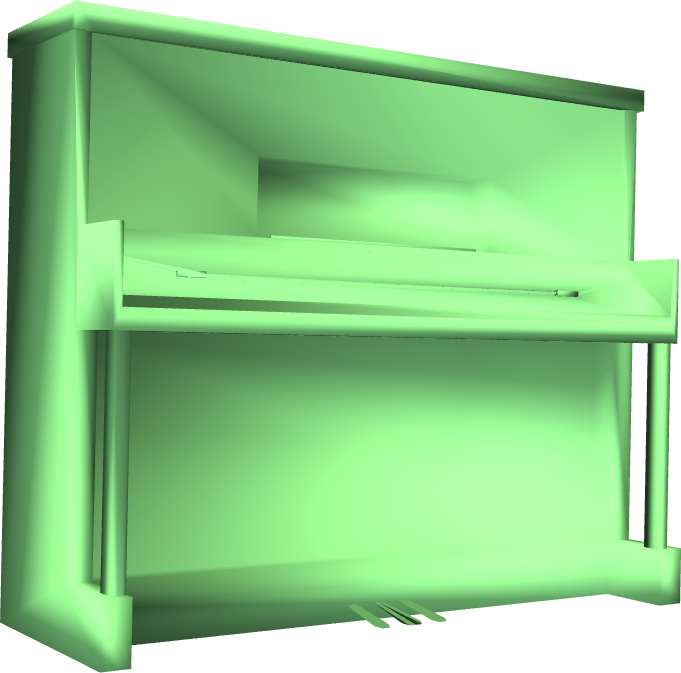}&
\includegraphics[width=0.17\textwidth]{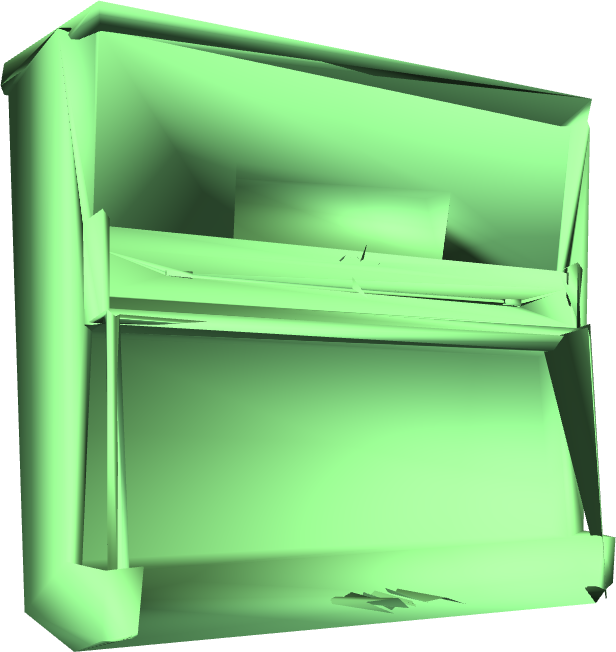}&
\includegraphics[width=0.17\textwidth]{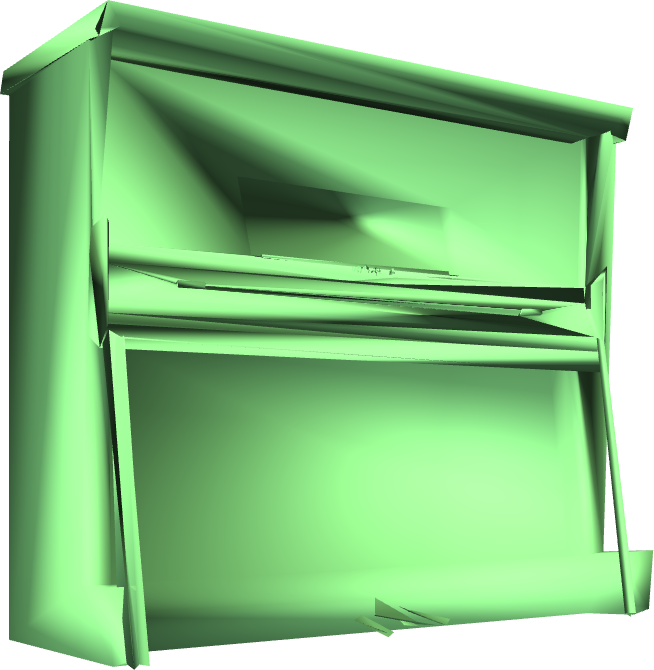} &
\includegraphics[width=0.055\textwidth]{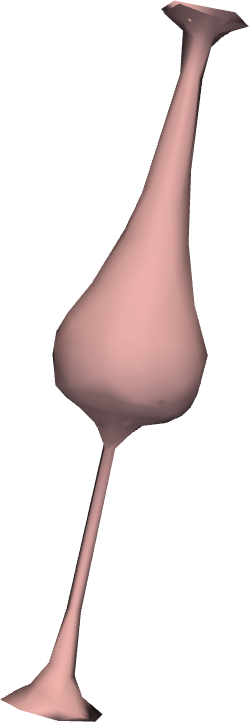}&
\includegraphics[width=0.055\textwidth]{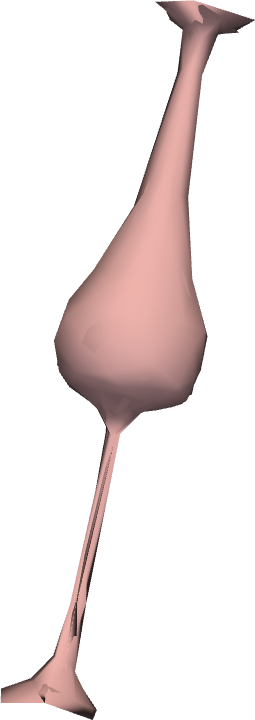}&
\includegraphics[width=0.0582\textwidth]{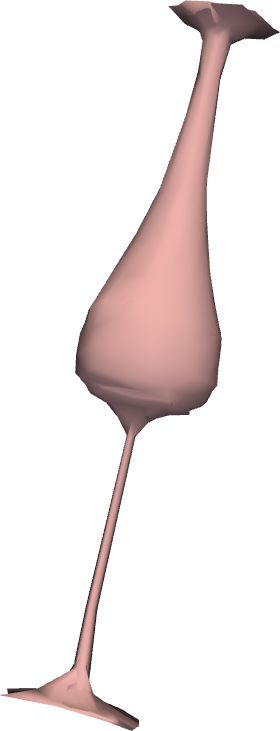}
\\
(a) Input &
(b) Walker attack
&
(c) MeshNet attack &
\hspace{-0.1in}(d) Input &
\hspace{-0.1in}(e) Walker 
&
\hspace{-0.1in}(f) MeshNet \\
Class: {\color{darkgreen}{Piano}} &
{\color{darkred}{Mantel}} &
{\color{darkred}{Tent}} &
{\color{darkgreen}{Vase}}&
{\color{darkred}{Flower pot}} &
{\color{darkred}{Glass box}} 
\end{tabular}
\caption{\label{fig:other_models_modelnet40}
{\bf Qualitative evaluation (ModelNet40).}
The attacked meshes seem to belong to the same class of their corresponding input meshes (in {\color{darkgreen}{green}}), yet they are misclassified  by the two networks that report results for ModelNet40~\cite{lahav2020meshwalker,feng2019meshnet} (misclassification in {\color{darkred}{red}}). 
}
\end{figure*}

\section{Experiments}
\label{sec:experiments}
Our proposed black-box attacks were tested on four SOTA mesh classification networks that differ significantly from each other: PD-meshNet~\cite{Milano20NeurIPS-PDMeshNet}, MeshCNN~\cite{hanocka2019meshcnn}, MeshNet~\cite{feng2019meshnet}, and MeshWalker~\cite{lahav2020meshwalker}.
Note that all SOTA classification networks are trained to be robust to uniform noise and sampling methods, and thus straightforward attacks will not be beneficial and educated attacks are necessary. 

For each given network, its imitating network was trained on (1)~the corresponding pre-processed  training mesh dataset and (2)~the prediction vectors of that dataset, queried from the attacked network.
The attack was preformed by applying Algorithm~\ref{alg:MeshWalkerTraining} on the test dataset.
Accuracy is defined as the percentage of correctly-predicted meshes,
measured by running the attacked test meshes through the original classification system.

The commonly-used datasets, SHREC11~\cite{veltkamp2011shrec} and ModelNet40~\cite{wu20153d}, were utilized.
They differ from one another in the number of classes, the number of objects per class, and the type of shapes they contain.
For each dataset we tested only the networks that report results on this dataset and present its reported results before the attacks.

\paragraph{SHREC11}
This dataset contains $600$ meshes, divided into $30$ classes, each containing $20$ models.
We follow the training setup of~\cite{ezuz2017gwcnn}, using a $16$/$4$-split per class.
The attacks are performed on three SOTA networks that report results on this dataset.
Table~\ref{tbl:accuracy}  compares the accuracy before and after the attack.
In all cases, the accuracy drops substantially, from $\geq 98.6\%$ to $\leq 18.3\%$.

\paragraph{ModelNet40}
This dataset contains $12,311$ CAD meshes, divided into $40$ categories of different sizes.
Out of the dataset,  $9,843$ meshes are used for training and $2,468$  are used for testing.
The models are not necessarily watertight and may contain multiple components, which is challenging for certain networks that require manifold data.
Therefore, we test our attacks only on the two recent networks that report results: MeshWalker~\cite{lahav2020meshwalker} and MeshNet ~\cite{feng2019meshnet}.
\begin{table}[tb]
\caption{
\label{tbl:accuracy}
{\bf Accuracy.} 
The accuracy of all networks drops substantially after the attacks, for all datasets.
}
\begin{center}
\begin{tabular}{|c|c|c|}
\hline
& {\bf SHREC11} & \\
\hline
Network & Pre-attack Accuracy & Post-attack Accuracy\\ 
\hline
 MeshWalker & $98.6\%$ & $16.0\%$ \\ 
\hline
 MeshCNN & $98.6\%$ & $14.8\%$ \\
\hline
 PD-MeshNet & $99.7\%$ & $18.3\%$ \\
\hline
\hline
& {\bf ModelNet40} & \\
\hline
Network & Pre-attack Accuracy & Post-attack Accuracy\\ 
\hline
 MeshWalker & $92.3\%$ & $10.1\%$ \\ 
\hline
MeshNet & $91.9\%$ & $12.0\%$ \\
\hline
\end{tabular}
\end{center}
\end{table}
Table~\ref{tbl:accuracy} compares the accuracy before  and after the attack.
In both cases, the accuracy drops significantly, from  $92.3\%$ and $91.9\%$ to $10.1\%$ and $12\%$, respectively.

\begin{figure}[tb]
\centering
\begin{tabular}{cccc}
\includegraphics[width=0.09\textwidth]{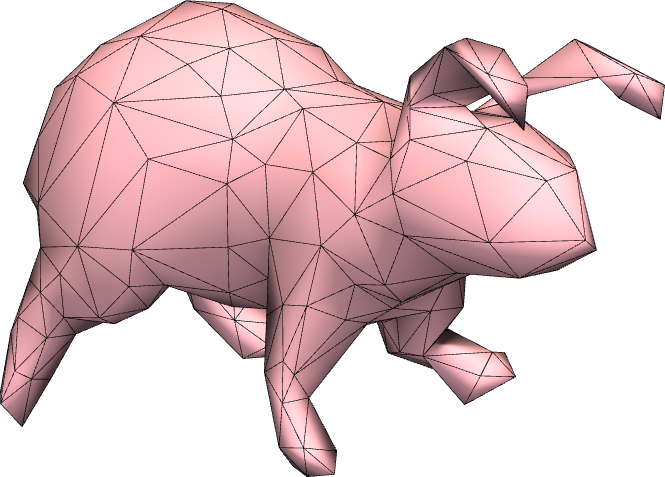}&
\includegraphics[width=0.09\textwidth]{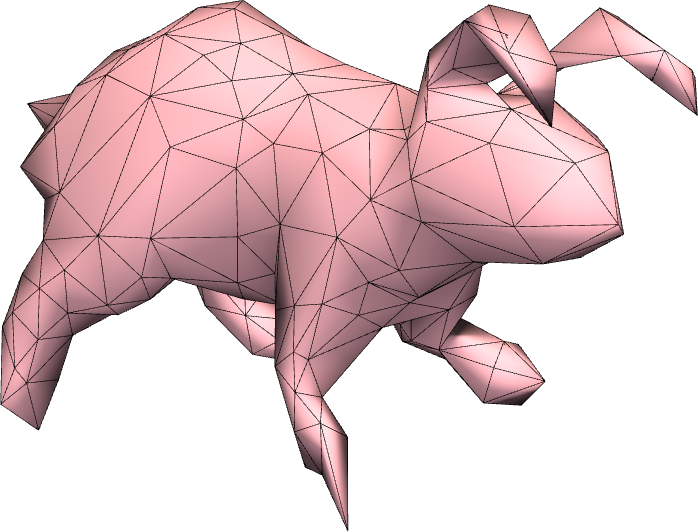}&
\includegraphics[width=0.09\textwidth]{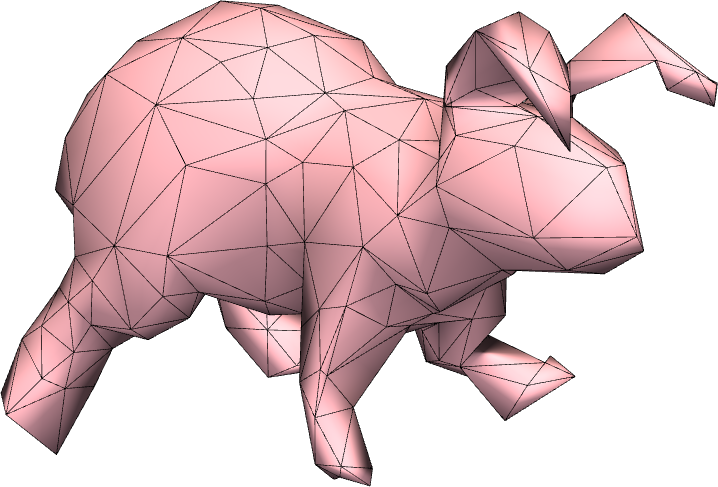}&
\includegraphics[width=0.09\textwidth]{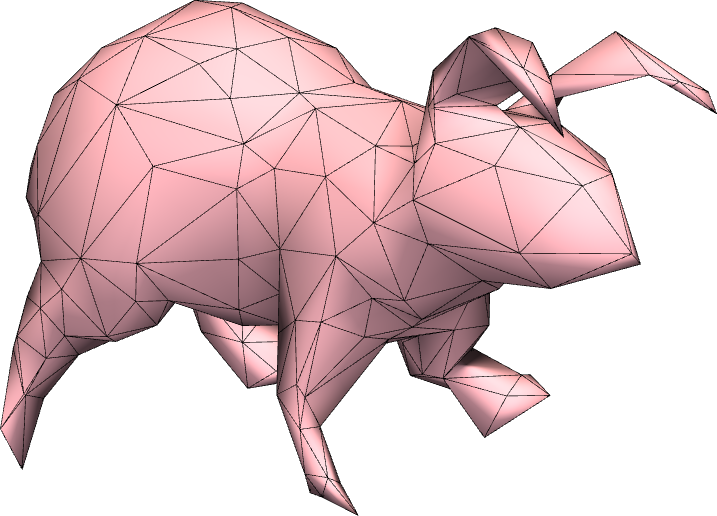}
\\
Class: {\color{darkgreen}{rabbit}} &
{\color{darkred}{man}} &
{\color{darkred}{armadillo}} &
{\color{darkred}{shark}} \\

\includegraphics[width=0.09\textwidth]{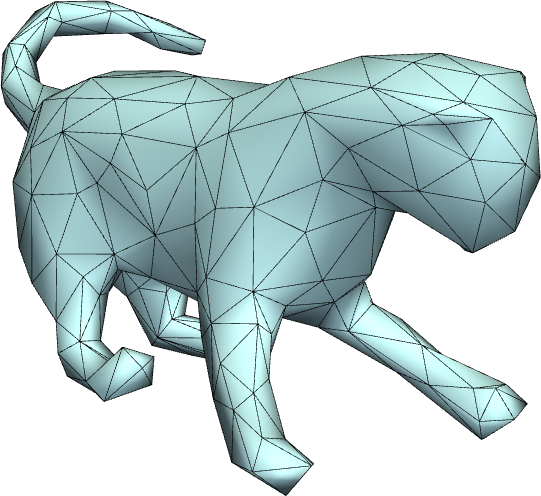}&
\includegraphics[width=0.09\textwidth]{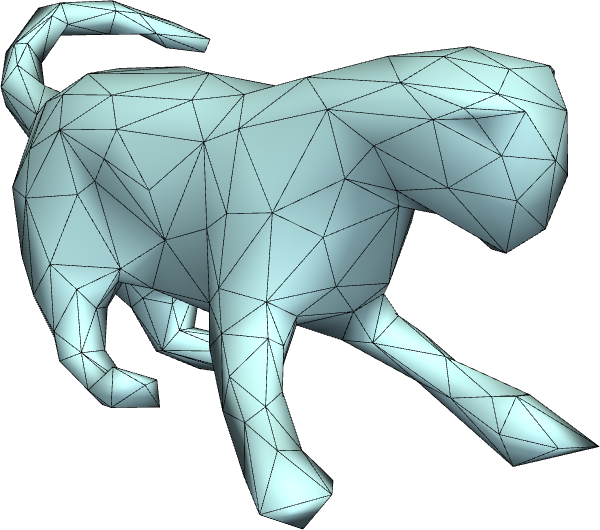}&
\includegraphics[width=0.09\textwidth]{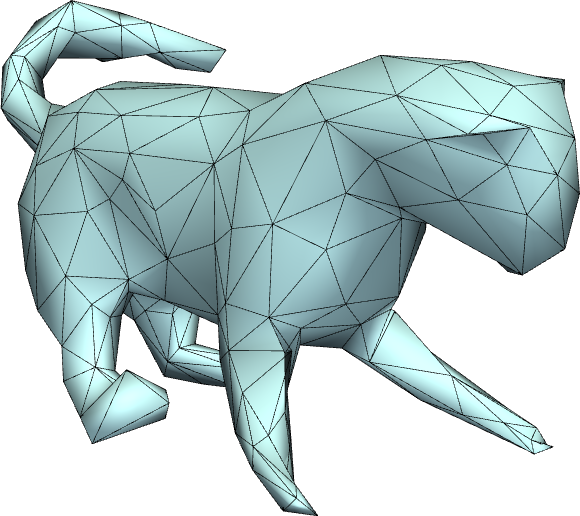}&
\includegraphics[width=0.09\textwidth]{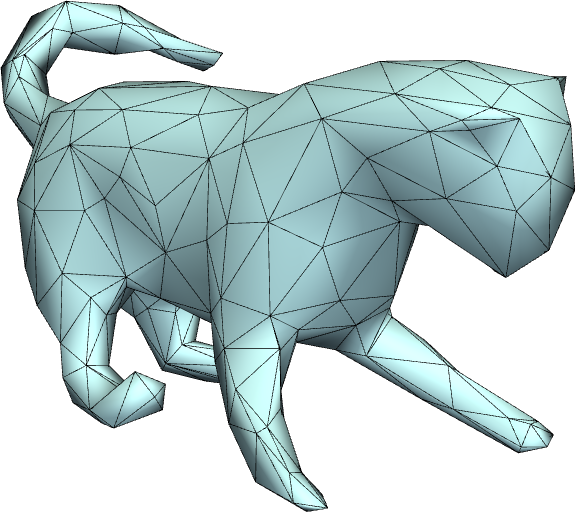}
\\
Class: {\color{darkgreen}{cat}} &
{\color{darkred}{rabbit}} &
{\color{darkred}{shark}} &
{\color{darkred}{bird}} \\

\includegraphics[width=0.09\textwidth]{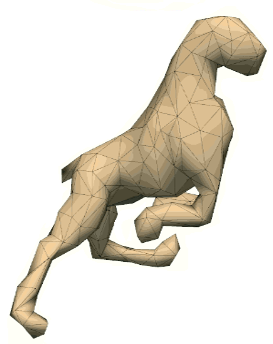}&
\includegraphics[width=0.09\textwidth]{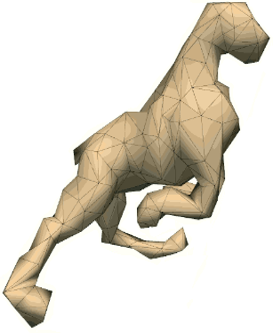}&
\includegraphics[width=0.09\textwidth]{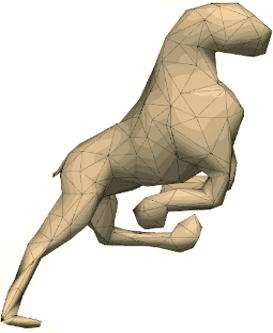}&
\includegraphics[width=0.09\textwidth]{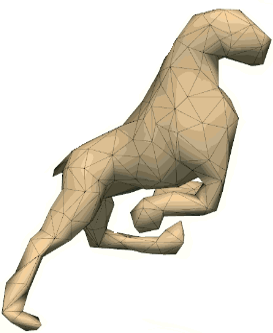}
\\
Class: {\color{darkgreen}{dog}} &
{\color{darkred}{horse}} &
{\color{darkred}{alien}} &
{\color{darkred}{shark}} \\
\includegraphics[width=0.09\textwidth]{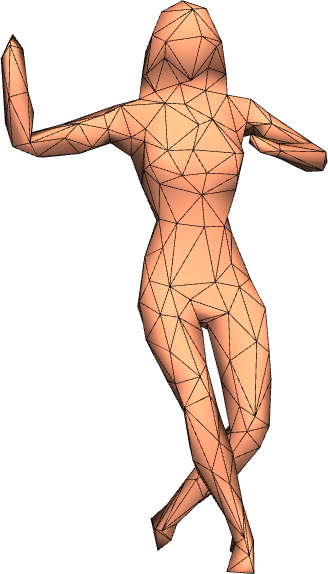}&
\includegraphics[width=0.09\textwidth]{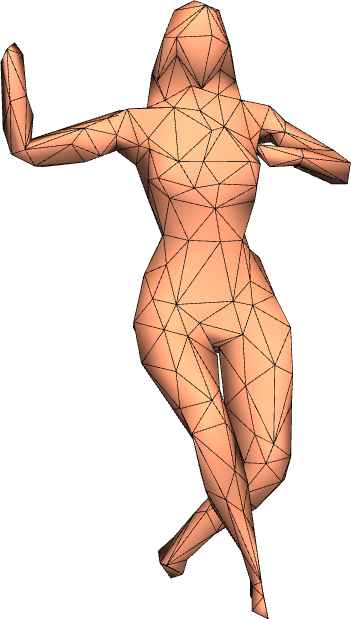}&
\includegraphics[width=0.09\textwidth]{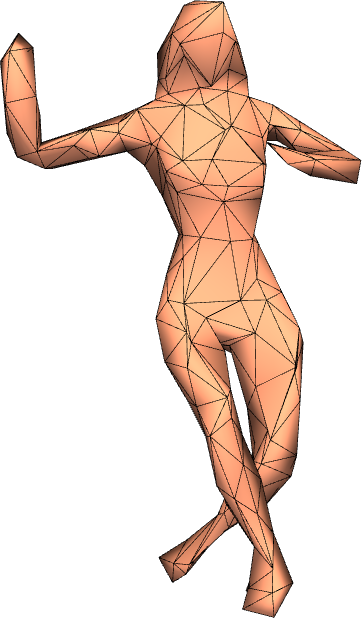}&
\includegraphics[width=0.09\textwidth]{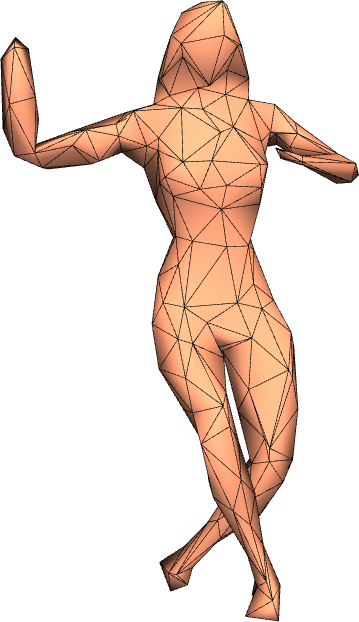}
\\
Class: {\color{darkgreen}{woman}} &
{\color{darkred}{gorilla}} &
{\color{darkred}{man}} &
{\color{darkred}{alien}} \\
(a) Input &
(b) MeshWalker 
&
\hspace{-0.05in}(c) PD-meshNet 
&
\hspace{-0.07in}(d) MeshCNN 
\end{tabular}
\caption{\label{fig:other_models_shrec11}
{\bf Qualitative evaluation (SHREC11).}
The attacked meshes look similar to the input meshes ({\color{darkgreen}{green}}) and seem to belong to the same class, but are misclassified by the networks trained on that dataset ({\color{darkred}{red}}). 
}
\end{figure}

Table~\ref{tbl:accuracy} verifies that we manage to cause misclassification.
Next, we verify, both quantitatively and qualitatively,  that the second requirement, (modifications will not result in misclassification by humans), is also satisfied.
Figures~\ref{fig:teaser}, \ref{fig:other_models_modelnet40} and \ref{fig:other_models_shrec11} show qualitative results, where the modifications are hardly visible.
This success is enhanced by the fact that we had to work on simplified meshes (as the networks work on them) and obviously, the fewer the vertices, the more difficult it is to move a vertex in an unnoticeable manner.

\begin{figure}[htb]
\centering
\includegraphics[width=0.48\textwidth]{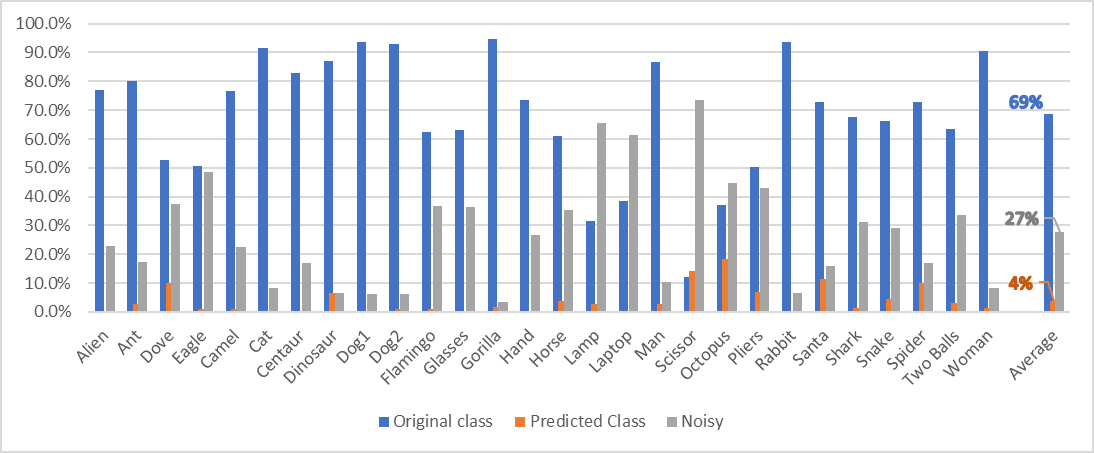}
\caption{
{\bf User Study.} 
For $69\%$ of the objects, the participants thought that the modified mesh and the input mesh belong to the same class.
When taking a closer look at the different classes, we note that our attacks are successful when applied to "natural" models, such as animals and humans.
However, man-made objects (e.g., scissors, lamps, laptops), which have straight geometric features, such as lines and corners, are harder to modify in an unnoticeable manner and they look noisy to the participants.
}
\label{fig:user_study_results}
\end{figure}

Furthermore, we held a user study on all the meshes of SHREC11, randomly choosing between the attacked mesh per input mesh. 
$79$~people participated in the user study, each saw $44$-$62$ meshes before and after the attack, such that each result was seen by $53$ people on average.
For each pair of (before, after) meshes, presented in a random order, the participants had to choose whether (1)~both meshes belong to the original class, (2)~one belongs to the original class and the other to the predicted class, or (3)~one mesh is noisy in comparison to the other.
As shown in Fig.~\ref{fig:user_study_results},  in $69\%$ of the cases, the participants thought that the modified mesh belongs to the same class of the input mesh.
In only  $4\%$  of the cases, the participants agreed with the misclassification.
However, in $27\%$ of the cases the participants found that the modified mesh is noisy; we will discuss this case below, in the limitations.

\begin{figure}[tb]
\centering
\begin{tabular}{ccc}
\includegraphics[width=0.09\textwidth]{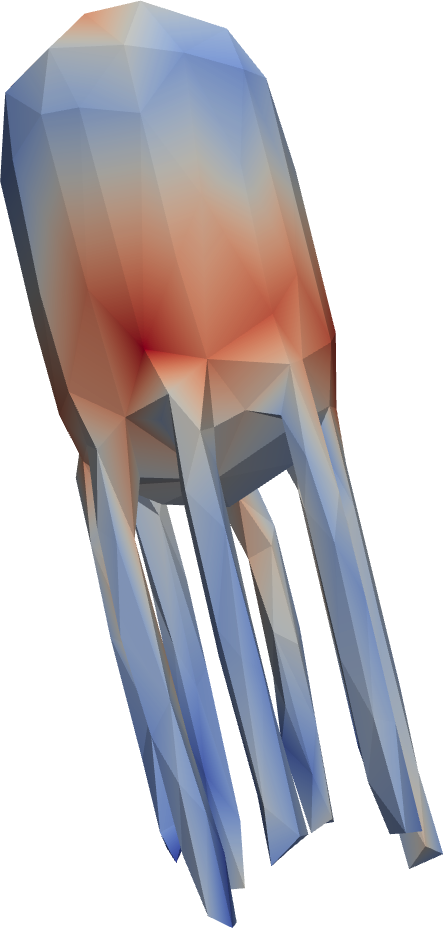} &
\includegraphics[width=0.09\textwidth]{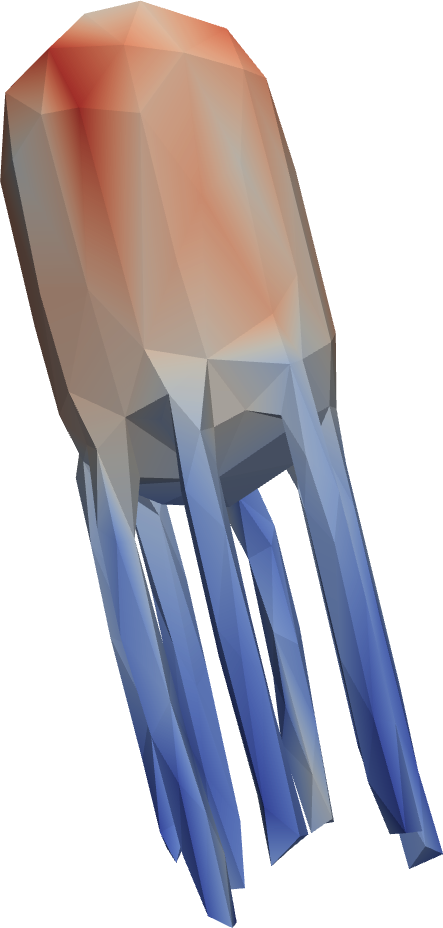} &
 \includegraphics[width=0.09\textwidth]{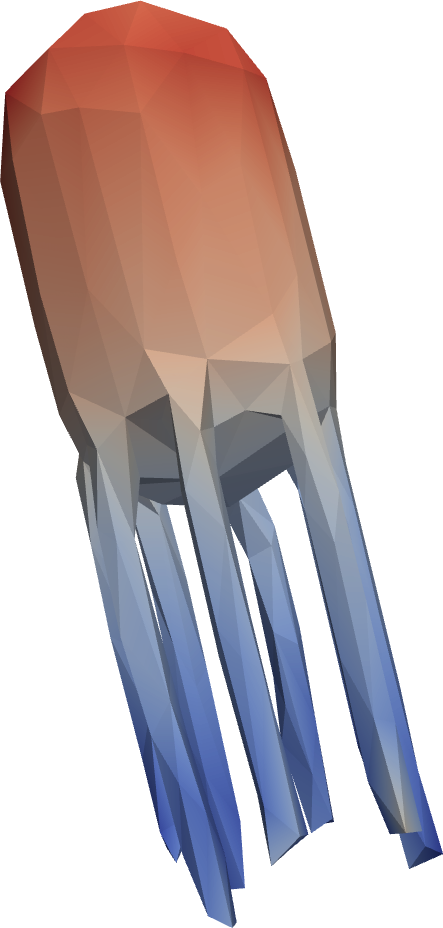} \\ 
    (a)  MeshWalker 
    & (b) PD-meshNet 
   & (c)  MeshCNN 
\end{tabular}
\caption{
{\bf Attack heat maps.} 
The imitating networks modify the octopus (from SHREC11)
differently for each attacked network.
The more reddish a vertex, the larger its modification.
}
\label{fig:heatmaps}
\end{figure}

Quantitatively, similarly to the case of images, we measured the distance between the input models and their modified versions.
In particular, we measured the $\mathcal{L}_2$ distance between the 3D coordinates of every vertex before and after the attack,
after normalizing each mesh to the unit sphere.
\begin{table}[tb]
\caption{{\bf Amount of modification.} 
 The $\mathcal{L}_2$ results on representative classes of SHREC11, in the descending order of the user study, demonstrate that all the networks require approximately the same amount of modification. 
 }
 \begin{center}
 \begin{tabular}{|c|c|c|c|}
 \hline
 Network & MeshCNN & PD-meshNet & MeshWalker \\ 
 \hline
 \hline
 Gorilla & $0.129$ & $0.116$ & $0.114$ \\
 \hline
 Cat & $0.123$ & $0.116$ & $0.114$\\
 \hline
 Man & $0.137$ & $0.131$ & $0.132$\\
 \hline
 Hand & $0.116$ & $0.114$ & $0.123$\\
 \hline
 Santa & $0.123$ & $0.123$ & $0.125$\\
 \hline
 Flamingo & $0.169$ & $0.162$ & $0.149$\\
 \hline
 Scissor & $0.147$ & $0.15$ & $0.151$\\
 \hline
 Average & $0.14$ & $0.15$ & $0.15$\\
 \hline
 \end{tabular}
 \label{tbl:classes}
 \end{center}
 \end{table}
Table~\ref{tbl:classes} shows representative results of several classes of SHREC11, listed in descending order of the user study results.
On average, the amount of modification is similar between all three networks.
However, there are differences when zooming into the classes.
For instance, MeshWalker applies larger modification to the Octopus class, MeshCNN to the Centaur class, and Pd-MeshNet to the Alien class.
This can be explained by the focus regions of the networks.
For example, for the octopus in Fig.~\ref{fig:heatmaps}, MeshWalker spreads the modifications across the whole object, rather than focusing on the head.  
As a result, in order to make a difference in the classification, this calls for bigger changes than those required by the other networks.

\paragraph{Comparison to white-box mesh adversarial networks}
For MeshNet, \cite{DBLP:journals/corr/abs-2104-12146} present a white-box adversarial network that attacks ModelNet40; it is designed for point clouds, but is  also tested on meshes.
Comparing our results to theirs, before the attack the accuracy of MeshNet~\cite{feng2019meshnet} is $91.9\%$.
After ~\cite{DBLP:journals/corr/abs-2104-12146}'s attack, the accuracy falls to  $25.5\%$, while after our attack it falls to $12.0\%$, i.e. our result is $13.5\%$ better.

\begin{figure}[tb]
\centering
\begin{tabular}{cccc}
\includegraphics[width=0.11\textwidth]{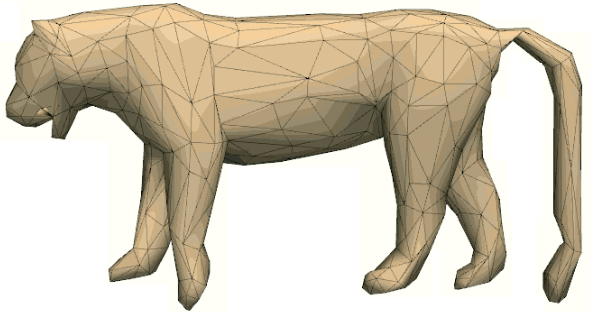}&
\includegraphics[width=0.11\textwidth]{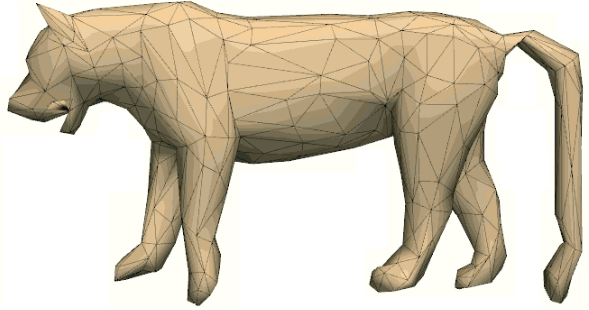}&
\includegraphics[width=0.09\textwidth]{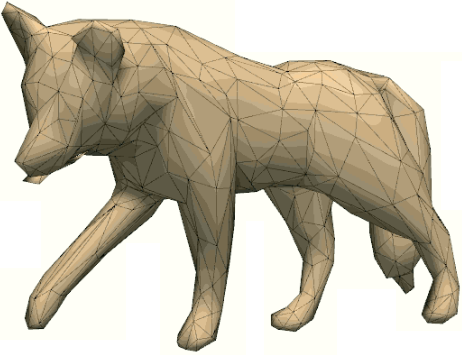}&
\includegraphics[width=0.09\textwidth]{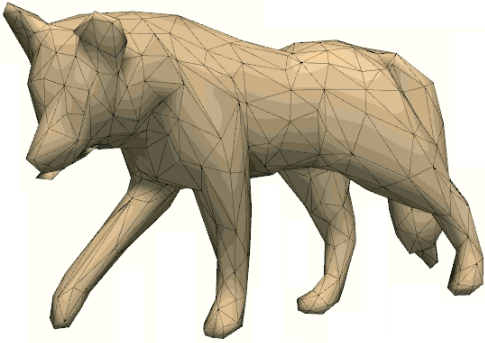}
\\
Class: {\color{darkgreen}{Big Cat}} &
{\color{darkred}{Horse}}&
{\color{darkgreen}{Dog}} &
{\color{darkred}{Hippo}} \\
\includegraphics[width=0.09\textwidth]{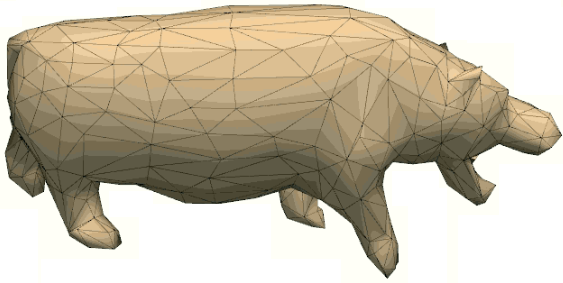}&
\includegraphics[width=0.09\textwidth]{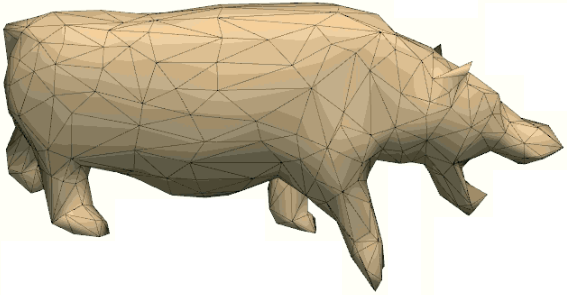}&
\includegraphics[width=0.09\textwidth]{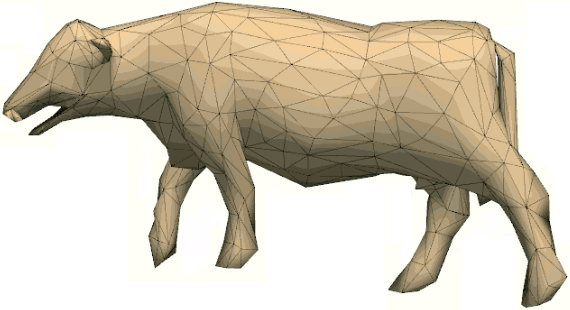}&
\includegraphics[width=0.09\textwidth]{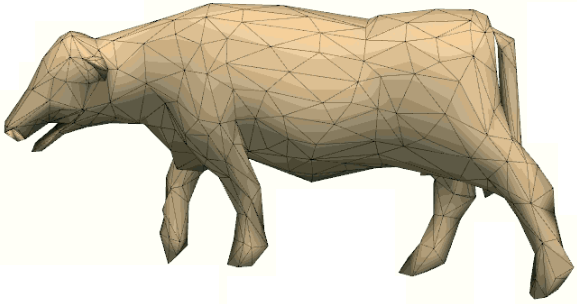}
\\
Class: {\color{darkgreen}{Hippo}} &
{\color{darkred}{Big Cat}}&
{\color{darkgreen}{Cow}} &
{\color{darkred}{Horse}} \\
(a) Input &
(b) Chebynet&
(a) Input &
(b) Chebynet
\end{tabular}
\caption{\label{fig:smal_models}
{\bf Qualitative evaluation (SMAL).}
The attacked meshes look similar to the input meshes ({\color{darkgreen}{green}}) and seem to belong to the same class, but are misclassified by the network trained on that dataset ({\color{darkred}{red}}).
}
\end{figure}

Mariani~\cite{mariani2020generating} and Rampini~\cite{rampini2021universal} present a white-box method to attack  ChebyNet~\cite{DBLP:journals/corr/DefferrardBV16}, which is a mesh classifier that was trained on the SMAL dataset~\cite{Zuffi:CVPR:2017}.
This dataset contains $600$ meshes, divided into $5$ varying-size classes.
We follow the training setup of~\cite{mariani2020generating} and~\cite{rampini2021universal} and use $480$ models for training and $120$ for testing.
Like them, we reach $100\%$ success rate;
The $\mathcal{L}_2$ distortions are $7.7e-2$ (\cite{mariani2020generating}), $3.6e-2$ (\cite{rampini2021universal}), and $3.6e-2$ (our method).
Fig.~\ref{fig:smal_models} shows some qualitative results, where it is evident that the barely-noticeable changes made by our method lead to misclassification of ChebyNet.

\section{Ablation study}
\label{sec:ablation}

\paragraph{On the importance of imitating networks.}
To understand the value of tailored imitating networks, Table \ref{tbl:Imitating_Networks} compares their results with those attained when providing a classification network with meshes attacked by an imitating network of a different classification network.
For instance, when attacking  MeshCNN with the imitating network of  MeshWalker,
the accuracy changed dramatically to  $51\%$, compared to  $14.8\%$ when attacked by its own  imitating system.
Thus, it is evident that the training process enables the networks to capture architecture-specific attributes.
Note that our imitating network is always vertex-based, whereas the imitated network may be based on other elements (e.g.,  MeshCNN is edge-based).

\begin{table}[tb]
\caption{
{\bf Imitating networks accuracy.} 
The best results are on the diagonal, i.e. each network
is best attacked by its own imitating network on the SHREC11 dataset.}
\begin{center}
\begin{tabular}{|c|c|c|c|}
\hline
imitating/attacked & 
MeshCNN & PD-meshNet & MeshWalker\\
\hline
\hline
MeshCNN 
& $14.8\%$ & $39.17\%$ & $49.16\%$ \\ 
\hline
PD-meshNet 
& $15\%$ & $18.3\%$ & $55.83\%$\\
\hline
MeshWalker 
& $51\%$ & $32.54\%$ & $16\%$\\
\hline
\end{tabular}
\label{tbl:Imitating_Networks}
\end{center}
\end{table}

This can be explained by looking at Fig.~\ref{fig:heatmaps}.
The heat-maps  show the amount of change in the different regions, i.e.
vertices colored red are modified more than vertices colored blue.
The areas that change the most are the ones that influence the classification the most.
We normalized the amount of modification for each mesh separately.
%
MeshCNN~\cite{hanocka2019meshcnn} attacks the tip of the octopus head more than other regions, PD-MeshNet~\cite{Milano20NeurIPS-PDMeshNet} spreads the modification differently across the head, whereas MeshWalker~\cite{lahav2020meshwalker} modifies the arms as well. 

Thus, our attacks provide a glimpse of how each system classifies and which areas are distinctive in "its eyes".
If indeed the imitating networks manage to learn the traits of networks, then we are one step closer to gaining insight onto how  different networks classify. 

\paragraph{On the choice MeshWalker.}
 Several factors led us to base our imitating network on MeshWalker~\cite{lahav2020meshwalker}:
First, since it is based on vertex features, it is possible to back-propagate the attack's modifications.
Second, it does not require the meshed to be a manifold, where some networks do  (~\cite{hanocka2019meshcnn}, \cite{Milano20NeurIPS-PDMeshNet}), making them prohibitive to Modelnet40~\cite{wu20153d}.
Finally, MeshWalker achieves SOTA results on both SHREC11~\cite{veltkamp2011shrec} and ModelNet40~\cite{wu20153d}, the two basic datasets for mesh classification.

For comparison, we used MeshNet~\cite{feng2019meshnet} as a white-box attacking  network, i.e. we attacked MeshNet itself, with knowledge of its gradients. 
We performed a similar procedure to that described in Algorithm~\ref{alg:MeshWalkerTraining}, on Modelnet40.
Specifically, for a given mesh we iteratively
(1) calculated the KLD loss between the prediction and the mesh's original class;
(2) extracted MeshNet's gradients;
(3) back-propagated the gradients all the way to the input layer and changed the mesh in the opposite direction of the gradients.
Since MeshNet features are face-based, during back-propagation the centers of the faces are the ones that  move. 
The coordinates of each vertex are interpolated between all the faces it is adjacent to.
The average {$\mathcal{L}_2$} distortion is $8\%$ larger than our results, when attempting to achieve the same percentage of misclassification.

\paragraph{Random perturbations.}
Our attack method entails well-educated changes. 
In order to ensure that the attacked networks are robust to non-educated random changes, we followed~\cite{lahav2020meshwalker, hanocka2019meshcnn, Milano20NeurIPS-PDMeshNet}'s perturbation scheme and randomly perturbed $30\%$ of the vertices in each mesh.
The accuracy dropped only by $0.1\%$ on MeshWalker \& MeshNet and by $0.3\%$ on PD-Meshnet \& MeshCNN.
This is not surprising, as most mesh classification networks are trained to be robust to random perturbations.

\begin{figure}[tb]
\centering
\begin{tabular}{cccc}
\includegraphics[width=0.08\textwidth]{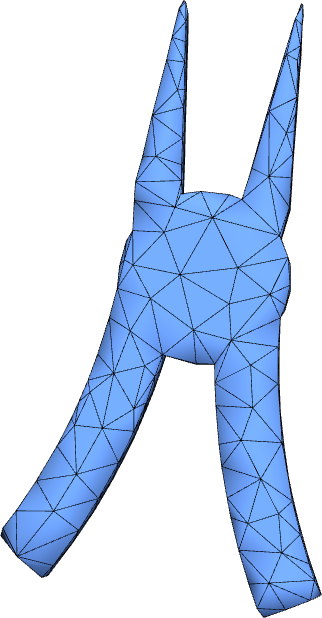}&
\includegraphics[width=0.08\textwidth]{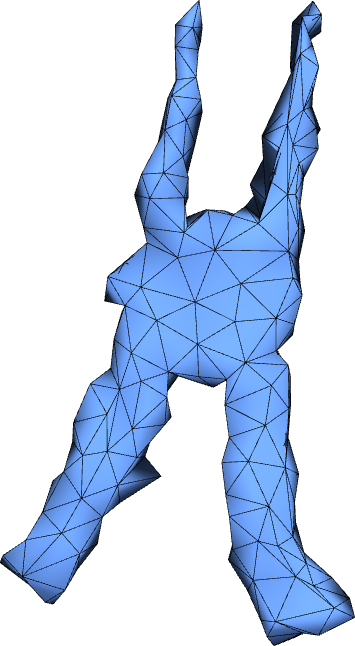}&
\includegraphics[width=0.075\textwidth]{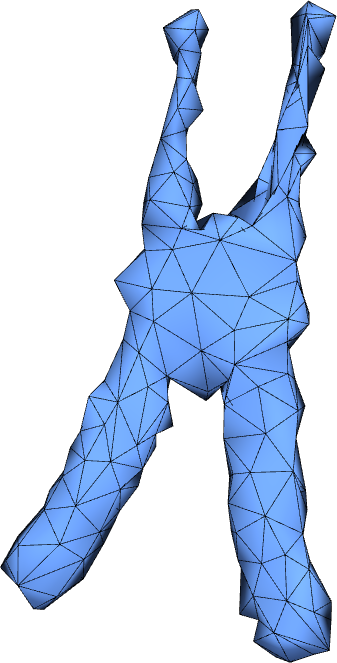}&
\includegraphics[width=0.075\textwidth]{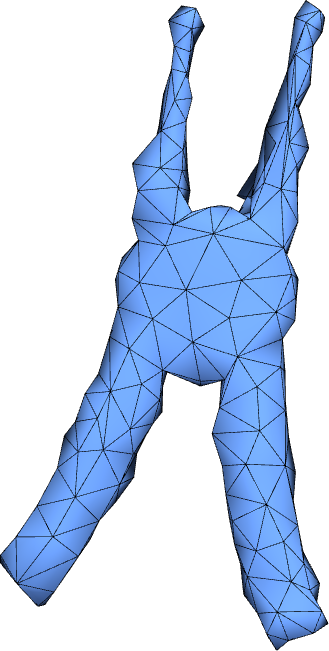}\\
(a) Input &
(b)  MeshWalker 
& (c) PD-meshNet 
&(d) MeshCNN 
\end{tabular}
\caption{
{\bf Limitations.} 
The attacked meshes might not maintain straight lines and corners.
Though the vertices moves by a small amount, the attacked meshes are considered too noisy by people, as indicated by our user study. 
}

\label{fig:limitations}
\end{figure}

\paragraph{Limitations.}
As indicated in Fig.~\ref{fig:user_study_results}, there are classes for which the modifications are considered too noisy by people, though the networks are being misled. 
This happens mostly for man-made objects, which have right dihedral angles and straight lines, whereas in the case of smooth natural objects it is less noticeable, such as humans and animals.
Fig.~\ref{fig:limitations} presents the worst case according to the user study, scissors from SHREC11.
We note that the noisiness of this class is not evident in Table~\ref{tbl:classes}. This is due to the fact that $\mathcal{L}_2$ cannot capture violation of geometric properties.

\section{Conclusion}

This paper has introduced a novel and unified black-box approach to adversarial attacks on mesh classification neural networks.
The key idea is to train an imitating network for each classification network we wish to attack.
This imitating network gets as input only the prediction vectors of the train dataset 
and manages to learn properties of the attacked network.
This is thanks to using a random walk-based network, which explores the mesh surface to its full.

Our network manages to change the meshes in a manner that causes networks to misclassify them, while the attacked meshes are  usually still classified correctly by people.
This is verified both quantitatively and qualitatively for four networks and on two datasets.
The mesh attack is performed by changing most of the vertices slightly, while moving the influential vertices a bit more.
As each attacked network finds different parts of the mesh more important for classification, our attacks shed some light on where these regions are.
This may be of assistance in better understanding the pitfalls of networks and improving their robustness.

In the future we would like to study targeted attacks.
Unlike untargeted attacks, where the goal is to simply cause misclassfication, in targeted attacks the aim is to cause the network to classify the attacked model as a specific class.
To start with, 
instead of calculating the $KLD$ between the prediction of the network and the source class one-hot vector in Algorithm~\ref{alg:MeshWalkerTraining}, it will be calculated between the prediction of the network and the target class one-hot vector.
Our preliminary study on a few classes from SHREC11~\cite{veltkamp2011shrec} achieved initial promising results.

The code is available at https://github.com/amirbelder/Random-Walks-for-Adversarial-Meshes.

\begin{acks}
We gratefully acknowledge the support of the Israel Science Foundation (ISF) 1083/18.
We thank Yizhak Ben-Shabat for his valuable comments during the rebuttal.
\end{acks}

\bibliographystyle{ACM-Reference-Format}
\bibliography{bibliography}

\end{document}